\documentclass[runningheads]{llncs}
\usepackage[T1]{fontenc}
\usepackage{xcolor}
\usepackage{multirow}          
\usepackage{booktabs}
\usepackage{graphicx,verbatim}
\usepackage{bbding}
\usepackage{caption}
\usepackage[colorlinks,
            linkcolor=blue,
            anchorcolor=blue,
            citecolor=blue]{hyperref}
\begin{document}
\title{Phenotype-Guided Generative Model for High-Fidelity Cardiac MRI Synthesis: Advancing Pretraining and Clinical Applications}
%

\author{Ziyu Li\inst{1} \and
Yujian Hu\inst{2} \and
Zhengyao Ding\inst{1} \and
Yiheng Mao\inst{1} \and
Haitao Li\inst{1} \and
Fan Yi\inst{1} \and
Hongkun Zhang\inst{2} \and
Zhengxing Huang\inst{1(}\Envelope\inst{)}
}

\authorrunning{Z. Li et al.}

\institute{Zhejiang University, China \email{\{liziyu,zhengyao.ding,yihengmao,lihaitao,fan\_yi,zhengxinghuang\}@zju.edu.cn}\and
The First Affiliated Hospital of Zhejiang University School of Medicine, China \email{\{huyujian,1198050\}@zju.edu.cn}}

\maketitle              
\begin{abstract}
Cardiac Magnetic Resonance (CMR) imaging is a vital non-invasive tool for diagnosing heart diseases and evaluating cardiac health. However, the limited availability of large-scale, high-quality CMR datasets poses a major challenge to the effective application of artificial intelligence (AI) in this domain. Even the amount of unlabeled data and the health status it covers are difficult to meet the needs of model pretraining, which hinders the performance of AI models on downstream tasks. In this study, we present Cardiac Phenotype-Guided CMR Generation (CPGG), a novel approach for generating diverse CMR data that covers a wide spectrum of cardiac health status. The CPGG framework consists of two stages: in the first stage, a generative model is trained using cardiac phenotypes derived from CMR data; in the second stage, a masked autoregressive diffusion model, conditioned on these phenotypes, generates high-fidelity CMR cine sequences that capture both structural and functional features of the heart in a fine-grained manner. We synthesized a massive amount of CMR to expand the pretraining data. Experimental results show that CPGG generates high-quality synthetic CMR data, significantly improving performance on various downstream tasks, including diagnosis and cardiac phenotypes prediction. These gains are demonstrated across both public and private datasets, highlighting the effectiveness of our approach. Code is availabel at https://anonymous.4open.science/r/CPGG.

\keywords{High-Fidelity CMR Generation \and Data Synthesis.}

\end{abstract}
\section{Introduction}
Cardiac Magnetic Resonance (CMR) imaging is a critical non-invasive diagnostic modality extensively employed for the diagnosis of cardiac diseases and the assessment of heart health\cite{bai2020population}\cite{wang2024screening}. The availability of large-scale, high-quality CMR datasets is essential for advancing the development and clinical implementation of artificial intelligence (AI) models in cardiac imaging. However, challenges related to data acquisition and privacy concerns have resulted in a scarcity of such datasets, thereby limiting the potential of AI in CMR image analysis. Even the amount of unlabeled data and the health status it covers are difficult to meet the needs of model pretraining, which further impedes the performance of AI models in clinical applications. To address these challenges, the application of generative AI for the creation of synthetic data have emerged as promising strategies to enhance dataset availability and model performance.

Recent studies have shown that generative AI models can effectively synthesize high-quality images in various medical imaging domains, such as cytopathological images \cite{shen2023cellgan}\cite{ye2023synthetic}, fundus images \cite{zhao2023label}, chest X-ray images \cite{bluethgen2024vision}\cite{hou2023diversity}, three-dimensional (3D) brain images \cite{tudosiu2024realistic}, and echocardiogram (ECHO) videos \cite{chen2024ultrasound}\cite{reynaud2023feature}, thus enhancing the understanding of medical images and improving downstream analyses \cite{wang2024self}. A common approach in these efforts involves using class labels or textual descriptions as conditions to control the image generation process. However, these conditions are inadequate for accurately describing the complex health status of the heart, hindering fine-grained control over the generation process to produce high-fidelity and diverse CMR data. Currently, there are only a few studies exploring the synthesis of CMR. \cite{al2023usability}\cite{zakeri2023dragnet} employ the first frame or mask as condition to synthesize CMR. However, the difficulty of obtaining these conditions on a large scale itself limits the feasibility of large-scale data generation. A closer parallel to our work lies in the synthesis of ECHO. \cite{reynaud2023feature} uses the Left Ventricle Ejection Fraction (LVEF) as a conditioning variable to generate echocardiogram videos. While this approach is valuable, the LVEF as a single condition remains relatively limited in capturing the full spectrum of cardiac health. Moreover, the direct application of a 3D diffusion framework for ECHO synthesis \cite{chen2024ultrasound}\cite{reynaud2023feature}\cite{zhou2024heartbeat} brings high training and inference costs, which hinder the generation of large-scale datasets necessary for model pretraining. These challenges underscore the need for more sophisticated and scalable methods in the generation of CMR images, where fine control over the synthesis of complex cardiac features is required.

Cardiac phenotypes encompasses a set of clinically relevant measurements extracted from CMR imaging, including key metrics such as LVEF and Left Ventricular End-Diastolic Volume (LVEDV), which together provide a comprehensive characterization of the heart's functional and structural properties. These phenotypes enable fine-grained, clinically interpretable control over the generation of CMR cine sequences, facilitating the creation of diverse, realistic samples reflecting various cardiac health status. In this work, \textbf{we propose a Cardiac Phenotype-Guided CMR Generation framework, which addresses the inherent complexity of high-dimensional CMR cine generation by decomposing the process into two distinct, manageable stages}.  In the first stage, we train a generative model that captures the underlying distribution of cardiac phenotypes, effectively representing CMR data into a low-dimensional space. In the second stage, we condition the CMR generation model on these cardiac phenotypes to generate CMR that exhibit diverse physiological characteristics. To efficiently synthesize large-scale, high-quality CMR data for pretraining purposes, \textbf{we adopt a masked autoregressive model for CMR generation}. \textbf{This approach offers a substantial speedup in inference} compared to traditional autoregressive video generation models and 3D diffusion models—which often require extensive autoregressive steps or larger computational resources. \textbf{We introduce diffusion loss to replace vector quantization, thereby enhancing responsiveness to fine-grained control}. Leveraging this framework, we generated a substantial volume of CMR data for pretraining and downstream tasks. Extensive experiments on the publicly available UK Biobank (UKB) dataset, and a private dataset across various tasks, consistently yielded state-of-the-art performance, proving the effectiveness of our approach.

\section{Method}
\begin{figure}
\includegraphics[width=\textwidth]{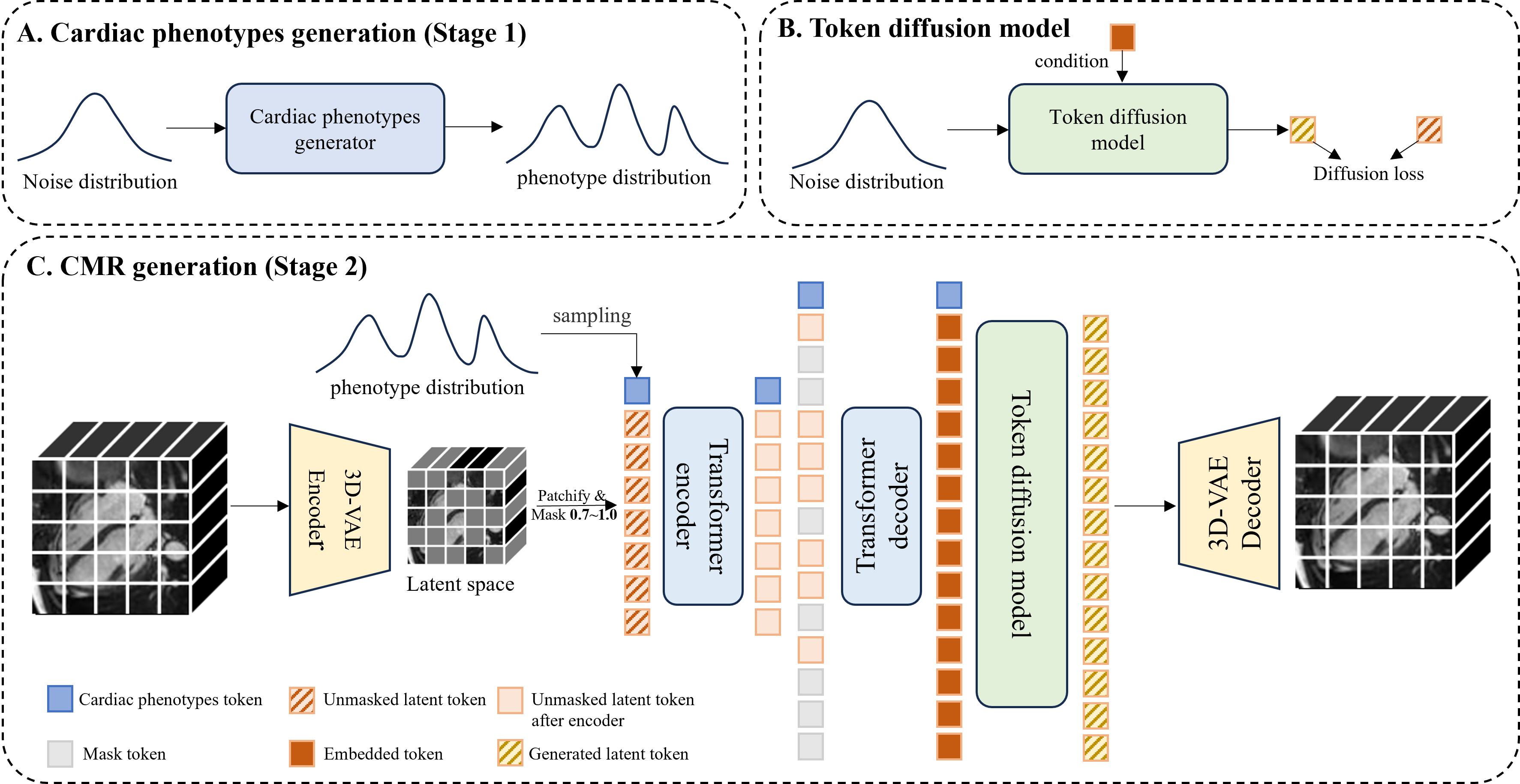}
\caption{Overview of our model. A and C describe a two-stage generation process. B showed the details of the generation of each token.} \label{fig1}
\end{figure}

Modeling the distribution of complex, high-dimensional cardiac MRI cines presents a significant challenge. We propose a novel Cardiac Phenotype-Guided CMR Generation (CPGG) framework, which decomposes the task of generating cardiac MRI data into two more manageable subtasks, as illustrated in Fig.~\ref{fig1}. 


\subsection{Cardiac Phenotypes Generation}
The cardiac phenotype is a compact, low-dimensional representation derived from CMR, encapsulating key functional and structural parameters of the heart. We use a generative model to capture their joint distribution. Our cardiac phenotypes generative model is based on the architecture of a Variational Autoencoder (VAE)\cite{kingma2013auto}, with stacked linear layers and LeakyReLU activation functions forming the network for both the encoder and decoder. During inference, a latent vector is sampled from a standard normal distribution and then decoded to generate cardiac phenotypes, effectively creating new phenotypes data. 

\subsection{Masked Autoregressive CMR Generative Model without Vector Quantization}
Video generation using diffusion models and autoregressive generative models has proven effective for synthesizing medical data, but the former is hindered by high training and inference costs, making it difficult to generate large volumes of data for pretraining and clinical applications, while the latter typically depend on well-trained discrete codebooks\cite{yu2023language}, thus when dealing with continuous fine-grained conditions such as cardiac phenotypes, this approach can lead to suboptimal performance during the generation process. Motivated by the work in \cite{li2024autoregressive}, we introduce a maskd autoregressive CMR generation model and eliminate the need for vector quantization, as depicted in Fig.~\ref{fig1}(c), to provide a faster and more refined generation method.

\noindent\textbf{3D-VAE} We modify the VAE in stable diffusion\cite{rombach2022high} to a 3D-VAE to project the input CMR  into a compressed latent space by extending 2D convolutions to 3D convolutions, the spatial downsampling factor, denoted as \( f_s \), and the temporal downsampling factor, denoted as \( f_t \). For an input CMR \( C \), with the shape \( 1 \times T \times H \times W \), the compressed latent representation \( x \) has the shape \( |x| \times \frac{T}{f_t} \times \frac{H}{f_s} \times \frac{W}{f_s} \), where \( |x| \) is the latent dimension.

\noindent\textbf{Masked Autoregressive Model with Diffusion Loss}
The Masked Autoregressive Model (MAM) is a variant of the standard autoregressive model. In contrast to traditional raster-order autoregression, MAM randomly predicts multiple tokens simultaneously based on the tokens that have already been observed that allowing for more flexible and efficient token generation. This autoregressive process can be formally expressed as:
\begin{equation}
P(X_1, \dots, X_K) = \prod_{k=1}^{K} P(X_k \mid X_1, \dots, X_{k-1}), \quad X_k = \{x_1, \dots, x_i\}
\end{equation}
where \( X_k \) represents the set of tokens predicted at the \( k \)-th step, \( x_i \) denotes an individual token. This design enables the parallel decoding of multiple tokens. We leverage Transformers as the backbone network for the masked autoregressive model, as bidirectional attention allowing all tokens to see each other that efficiently model spatiotemporal dependencies. Our implementation follows a framework similar to the Masked Autoencoder\cite{he2022masked} (MAE). Specifically, for a CMR compressed representation generated by a 3D-VAE encoder, we partition the representation into non-overlapping tokens \( x_i \), each with a shape of \( |x| \times p_t \times p_s \times p_s \), where \( p_t \) denotes the temporal stride and \( p_s \) represents the spatial stride. These tokens are randomly masked, during training, a dynamic masking ratio is applied, as used in \cite{li2023mage}\cite{li2024autoregressive}.

We further incorporate a diffusion model to capture the distribution of each token in continuous space, which we refer to as the token diffusion model. Let \( x_i \) represents a ground-truth token, and \( z_i \) denote the vector embedded by the autoregressive model at the corresponding position, i.e., \( z_i = f(X_1, \dots, X_{k-1}) \). The objective is to model the distribution of \( x_i \) conditioned on \( z_i \), that is, \( p(x_i|z_i) \). We introduce the diffusion loss\cite{li2024autoregressive} to model this conditional distribution:
\begin{equation}
\mathcal{L}(z_i, x_i) = {E}_{\varepsilon, t} \left[ \left\| \varepsilon - \varepsilon_\theta(x_i^t \mid t, z_i) \right\|^2 \right].
\end{equation}
where \( \varepsilon \) denotes the noise vector sampled from \( \mathcal{N}(0, I) \), and \( x_i^t \) represents the noisy token at the time step \(t\), defined as \( x_i^t = \sqrt{\bar{\alpha}_t} x_i + \sqrt{1 - \bar{\alpha}_t} \, \varepsilon\), where \( \bar{\alpha}_t \) is a noise schedule. The noise estimator \(\varepsilon_\theta\) is implemented using a small MLP network. The loss is only computed for masked tokens. During inference, we iteratively sample \( x_i^T \) to \( x_i^0 \) via a reverse diffusion procedure, defined as:
\begin{equation}
x_i^{t-1} = \frac{1}{\sqrt{\alpha_t}} \left( x_i^t - \frac{1 - \alpha_t}{\sqrt{1 - \bar{\alpha}_t}} \varepsilon_\theta(x_i^t \mid t, z_i) \right) + \sigma_t \delta.
\end{equation}
where \( x_i^T \) and \(\delta\) are sampled from \( \mathcal{N}(0, I) \), \(\sigma_t\) is the noise level at time step \(t\).

\noindent\textbf{Cardiac Phenotypes Conditioning}
The cardiac phenotypes vector is projected into the token dimension through an MLP network, after which it is concatenated to the start of the encoded sequence as the [CLS] token. This input sequence is then processed using bidirectional attention, allowing each token to incorporate the conditional information from the cardiac phenotypes. 

\noindent\textbf{Iterative Decoding}
We generate CMR cine sequences using an iterative decoding strategy like the approach outlined in \cite{li2023mage}. The process begins with an empty CMR latent representation, where all tokens are masked. The iterative decoding proceeds over \(K\) steps, during which the model predicts the remaining masked tokens at each iteration, and the predicted tokens are randomly retained, masking ratio adhering to a cosine schedule. This ensures that the model progressively refines the CMR representation across iterations.

\section{Experiments}
\subsection{Datasets and Experiments Setting}
In this study, we utilized four-chamber CMR cine sequences from the first imaging assessment of the UK Biobank to construct datasets. A total of 32,444 CMR cine sequences were screened, with 82 cardiac phenotypes being complete and available for CMR generation and cardiac phenotypes prediction. The dataset was divided into 25,955 samples for the training set, 3,244 samples for the validation set, and 3,245 samples for the test set. For the disease classification task, we respectively selected 196, 5,464 and 578 participants to construct three datasets: cardiomyopathy (UKB-CM), coronary artery disease (UKB-CAD) and heart failure (UKB-HF), ensuring a 1:1 ratio of positive to negative cases. Additionally, we collected a separate cardiomyopathy dataset (CMDS) from hospital A for tasks involving cardiomyopathy diagnosis (binary classification) and cardiomyopathy subtype classification (four-class classification). This dataset comprises 535 samples, including 195 cases of hypertrophic cardiomyopathy, 160 cases of dilated cardiomyopathy, 33 cases of restrictive cardiomyopathy, and 147 healthy controls. We use five-fold cross-validation for disease classification tasks.

For each dataset, we employed a segmentation model from \cite{bai2020population} to extract the heart region and resized  to $1\times50\times96\times 96$.
For the 3D-VAE, we set the latent dimension $|x|$ to 16, $f_t$ to 2, and $f_s$ to 8. In the masked autoregressive model, we implemented a 12-layers encoder and decoder, with a latent dimension of 768 and a patch size of 5×2×2. The patch size was chosen to reduce computation, as CMR data typically exhibit significant redundancy in the temporal dimension. During training, we randomly sampled a mask ratio between 0.7 and 1.0. In the diffusion process, we used a cosine-shaped noise schedule with 1000 steps during training and 100 steps during inference. The denoising MLP was constructed with 3 blocks and a width of 1024 channels. In the iterative decoding process, we performed 16 steps to progressively generate CMR data.
We implemented an MAE-base\cite{he2022masked} framework for pretraining, using a mask ratio of 0.75, with a patch size of $16 \times 16$, the temporal dimension of CMR data is treated as the channel dimension\cite{ding2024cross}.
All experiments were performed on an NVIDIA A800 GPU. The AdamW optimizer was employed for training, with a learning rate of $8e-4$ for the CMR generative model, $5e-4$ for model pretraining, and $5e-5$ for finetuning on downstream tasks. The generative and pretraining models were trained for 400 epochs, while finetuned models underwent training for 100 epochs.

\begin{table}[!b]
\centering
\caption{Quantitatively evaluation of our CPGG model. The inference time is the average time to generate each CMR with a batch size of 16 using one A800 GPU. CFG means classifier-free guidance.}
\label{gen_table}
\renewcommand{\arraystretch}{1.1} 
\resizebox{0.7\textwidth}{!}{
\setlength{\tabcolsep}{9pt}
\begin{tabular}{cccc} 
\toprule
Method                 & FID$\downarrow$ & FVD$\downarrow$ & Inference time                    \\ 
\hline
VideoGPT\cite{yan2021videogpt}(uncond)       & 40.58           & 1320.26         & \multirow{2}{*}{1.87 sec / vid.}  \\
VideoGPT(CFG=3.0)      & 36.32           & 1234.46         &                                   \\ 
\hline
ModelScopeT2V\cite{wang2023modelscope}(uncond)  & 50.86           & 1515.44        & \multirow{2}{*}{4.26 sec / vid.}  \\
ModelScopeT2V(CFG=3.0) & 26.86           & 961.80         &                                   \\ 
\hline
CPGG(uncond)           & 19.85           & 760.63          & \multirow{2}{*}{0.36 sec / vid.}  \\
CPGG(CFG=3.0)          & \textcolor{red}{15.14}  & \textcolor{red}{711.17} &                                   \\
\bottomrule
\end{tabular}
}
\end{table}

\subsection{CMR Generation Quality}
\begin{figure}[htbp]
\centering
\includegraphics[width=\textwidth]{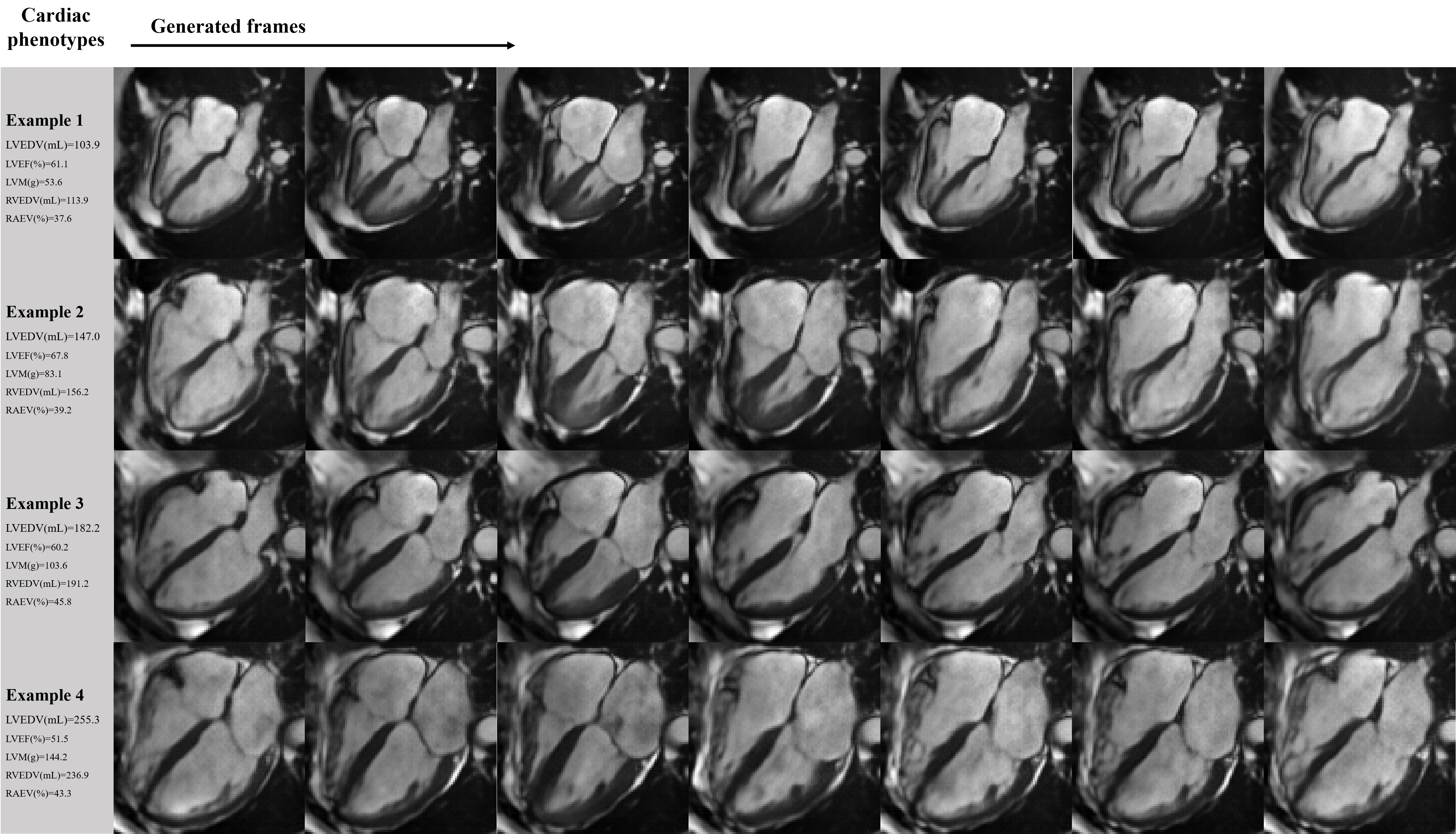}
\caption{Examples of generated CMR and their corresponding cardiac phenpotypes using the CPGG framework, ordered by LVEDV from small to large.} \label{fig2}
\end{figure}

We compare our approach with the previous state-of-the-art autoregressive generative model using vector quantization (VideoGPT) and 3D diffusion model (ModelScopeT2V). As shown in Table \ref{gen_table}, our model achieves better performance on both Fréchet Inception Distance\cite{heusel2017gans} (FID) and Fréchet Video Distance\cite{unterthiner2019fvd} (FVD) metrics. We also observe that applying the proposed cardiac phenotype-guided conditioning to each method improves generation quality, and our approach achieving the best performance. Furthermore, our model demonstrates a significant speedup in inference compared to other models, making it possible to synthesize large amounts of data for augmenting pretraining data. Figure \ref{fig2} presents examples of generated CMR and their corresponding cardiac phenotypes, ordered by LVEDV from small to large. It can be intuitively observed that the generated CMR can respond well to fine-grained control of cardiac phenotypes and exhibits high fidelity.

\subsection{Performance of Downstream Tasks after Data Mixing}
\noindent\textbf{Disease Classicifation} To further explore the usability of synthetic data, we used the generated CMR to augment the pretraining data and compared performance on the disease classification task. As shown in Table \ref{cls_table_ukb} and Table \ref{cls_table_zju}, as the proportion of synthetic data gradually increases, we observe a corresponding improvement in classification performance across all datasets. When the amount of synthetic data reaches five times that of the real data, a significant performance improvement is achieved compared to pretraining with real data alone in all datasets. Therefore, our method can be seen as an effective data augmentation strategy that benefits both pretraining and the disease classification tasks.

\noindent\textbf{Cardiac Phenotypes Prediction} Furthermore, we conducted the cardiac phenotypes prediction task, the performance of several important phenotypes and the average $R^2$ across 82 cardiac phenotypes prediction are reported in Table \ref{cls_table_ukb} and Fig \ref{fig3}. When only using synthetic data to augment the pretraining data, we observed a gradual improvement in prediction performance. Building upon this, incorporating the synthetic phenotypes-CMR data into the training set for the cardiac phenotypes prediction task further improved the average $R^2$. This proves that the CMR data generated by our method has high fidelity and strictly adheres to fine-grained conditions such as cardiac phenotypes.

\begin{table}[!h]
\centering
\renewcommand{\arraystretch}{1.1} 
\setlength{\tabcolsep}{8pt} 
\caption{Performance of two downstream tasks on public datasets. The upper part of the table is the disease classification task, and the lower part is the cardiac phenotype regression task. * indicates that synthetic CMR data is used not only for data augmentation in the pretraining stage but also as labeled data for data augmentation during finetuning.}
\label{cls_table_ukb}
\resizebox{\textwidth}{!}{
\begin{tabular}{ccllcllccllcllc}
\toprule
Disease                                       & \multicolumn{5}{c}{UKB-CAD}                                       & \multicolumn{4}{c}{UKB-CM}                                        & \multicolumn{5}{c}{UKB-HF}                                         \\
\hline
Method                                        & \multicolumn{2}{c}{ACC}         & \multicolumn{3}{c}{AUC}         & \multicolumn{2}{c}{ACC}         & \multicolumn{2}{c}{AUC}         & \multicolumn{3}{c}{ACC}         & \multicolumn{2}{c}{AUC}          \\
\hline
ViT\cite{alexey2020image}    & \multicolumn{2}{c}{0.681±0.005} & \multicolumn{3}{c}{0.741±0.014} & \multicolumn{2}{c}{0.730±0.018} & \multicolumn{2}{c}{0.778±0.064} & \multicolumn{3}{c}{0.732±0.015} & \multicolumn{2}{c}{0.800±0.026}  \\
MAE\cite{he2022masked}(real) & \multicolumn{2}{c}{0.719±0.156} & \multicolumn{3}{c}{0.787±0.022} & \multicolumn{2}{c}{0.816±0.057} & \multicolumn{2}{c}{0.843±0.067} & \multicolumn{3}{c}{0.799±0.026} & \multicolumn{2}{c}{0.881±0.008}  \\
–mix   100\%                                  & \multicolumn{2}{c}{0.726±0.078} & \multicolumn{3}{c}{0.804±0.012} & \multicolumn{2}{c}{0.821±0.046} & \multicolumn{2}{c}{0.863±0.041} & \multicolumn{3}{c}{0.836±0.017} & \multicolumn{2}{c}{0.892±0.013}  \\
–mix   200\%                                  & \multicolumn{2}{c}{0.730±0.013} & \multicolumn{3}{c}{0.807±0.011} & \multicolumn{2}{c}{0.821±0.020} & \multicolumn{2}{c}{0.867±0.047} & \multicolumn{3}{c}{0.829±0.013} & \multicolumn{2}{c}{0.904±0.019}  \\
–mix   300\%                                  & \multicolumn{2}{c}{0.731±0.012} & \multicolumn{3}{c}{0.809±0.015} & \multicolumn{2}{c}{0.831±0.035} & \multicolumn{2}{c}{0.876±0.041} & \multicolumn{3}{c}{0.834±0.034} & \multicolumn{2}{c}{0.900±0.020}  \\
–mix   400\%                                  & \multicolumn{2}{c}{0.735±0.011} & \multicolumn{3}{c}{0.811±0.008} & \multicolumn{2}{c}{0.831±0.054} & \multicolumn{2}{c}{0.875±0.044} & \multicolumn{3}{c}{\textcolor{red}{0.848±0.041}} & \multicolumn{2}{c}{0.910±0.022}  \\
–mix   500\%                                  & \multicolumn{2}{c}{\textcolor{red}{0.739±0.014}} & \multicolumn{3}{c}{\textcolor{red}{0.812±0.012}} & \multicolumn{2}{c}{\textcolor{red}{0.841±0.043}} & \multicolumn{2}{c}{\textcolor{red}{0.878±0.036}} & \multicolumn{3}{c}{0.844±0.038} & \multicolumn{2}{c}{\textcolor{red}{0.913±0.031}}  \\
\toprule
Phenotype                                     & \multicolumn{3}{c}{LVEDV(mL)}      & \multicolumn{3}{c}{LVEF(\%)}      & \multicolumn{2}{c}{LVM(g)}         & \multicolumn{3}{c}{RVEDV(mL)}      & \multicolumn{3}{c}{RAEF(\%)}                            \\
\hline
Method                                        & MAE    & \multicolumn{2}{c}{$R^2$} & MAE   & \multicolumn{2}{c}{$R^2$} & MAE                        & $R^2$ & \multicolumn{2}{c}{MAE}    & $R^2$ & \multicolumn{2}{c}{MAE}   & $R^2$                       \\
\hline
ViT\cite{alexey2020image}    & 11.381 & \multicolumn{2}{c}{0.782} & 3.490  & \multicolumn{2}{c}{0.470}  & 6.352                      & 0.839 & \multicolumn{2}{c}{12.510}  & 0.794 & \multicolumn{2}{c}{5.078} & 0.484                       \\
MAE\cite{he2022masked}(real) & 10.826 & \multicolumn{2}{c}{0.806} & 3.553 & \multicolumn{2}{c}{0.457} & 6.479                      & 0.836 & \multicolumn{2}{c}{12.048} & 0.808 & \multicolumn{2}{c}{5.102} & 0.474                       \\
–mix   100\%                                  & 10.120 & \multicolumn{2}{c}{0.827} & 3.432 & \multicolumn{2}{c}{0.496} & 5.929                      & 0.868 & \multicolumn{2}{c}{11.589} & 0.822 & \multicolumn{2}{c}{4.930} & 0.509                       \\
–mix   200\%                                  & 10.200 & \multicolumn{2}{c}{0.826} & 3.396 & \multicolumn{2}{c}{0.503} & 5.568                      & 0.876 & \multicolumn{2}{c}{11.368} & 0.830 & \multicolumn{2}{c}{4.816} & 0.527                       \\
–mix   300\%                                  & 9.784  & \multicolumn{2}{c}{0.843} & 3.410 & \multicolumn{2}{c}{0.502} & 5.678                      & 0.875 & \multicolumn{2}{c}{11.057} & 0.837 & \multicolumn{2}{c}{4.762} & 0.542                       \\
–mix   400\%                                  & 9.721  & \multicolumn{2}{c}{0.844} & 3.328 & \multicolumn{2}{c}{0.523} & 5.540                      & 0.883 & \multicolumn{2}{c}{11.201} & 0.837 & \multicolumn{2}{c}{4.544} & 0.577                       \\
–mix   500\%                                  & 9.776  & \multicolumn{2}{c}{0.844} & 3.354 & \multicolumn{2}{c}{0.513} & 5.599                      & 0.882 & \multicolumn{2}{c}{11.145} & 0.838 & \multicolumn{2}{c}{4.660} & 0.558                       \\
\hline
–mix*   100\%                                 & 10.081 & \multicolumn{2}{c}{0.830}  & 3.426 & \multicolumn{2}{c}{0.500} & 5.835                      & 0.871 & \multicolumn{2}{c}{11.652} & 0.820 & \multicolumn{2}{c}{4.994} & 0.496                       \\
–mix*   200\%                                 & 10.083 & \multicolumn{2}{c}{0.829} & 3.391 & \multicolumn{2}{c}{0.507} & 5.684                      & 0.875 & \multicolumn{2}{c}{11.458} & 0.831 & \multicolumn{2}{c}{4.800} & 0.528                       \\
–mix*   300\%                                 & \textcolor{red}{9.629}  & \multicolumn{2}{c}{\textcolor{red}{0.846}} & 3.383 & \multicolumn{2}{c}{0.505} & 5.431                      & 0.884 & \multicolumn{2}{c}{11.153} & 0.833 & \multicolumn{2}{c}{4.743} & 0.546                       \\
–mix*   400\%                                 & 9.713  & \multicolumn{2}{c}{0.844} & 3.337 & \multicolumn{2}{c}{0.504} & \textcolor{red}{5.322}                      & \textcolor{red}{0.888} & \multicolumn{2}{c}{11.224} & 0.834 & \multicolumn{2}{c}{\textcolor{red}{4.478}} & \textcolor{red}{0.587}                       \\
–mix*   500\%                                 & 9.639  & \multicolumn{2}{c}{0.842} & \textcolor{red}{3.327} & \multicolumn{2}{c}{\textcolor{red}{0.525}} & 5.621                      & 0.875 & \multicolumn{2}{c}{\textcolor{red}{11.016}} & \textcolor{red}{0.840} & \multicolumn{2}{c}{4.529} & 0.584                      \\
\bottomrule
\end{tabular}
}
\end{table}

\begin{figure}[!h]
    \centering
    \begin{minipage}{0.56\textwidth}
        \centering
        \small 
        \renewcommand{\arraystretch}{1.1} 
        \setlength{\tabcolsep}{4pt} 
         \captionof{table}{Performance of disease classification task on private dataset.}
        \label{cls_table_zju}
        \resizebox{\textwidth}{!}{ 
            \begin{tabular}{ccccc} 
                \toprule
                Dataset    & \multicolumn{2}{c}{CMDS\(^b\)}                 & \multicolumn{2}{c}{CMDS\(^f\)}                  \\ 
                \hline
                Method     & ACC                  & AUC                  & ACC                  & AUC                   \\ 
                \hline
                ViT\cite{alexey2020image}        & 0.759±0.025          & 0.696±0.055          & 0.555±0.017          & 0.718±0.061           \\
                MAE\cite{he2022masked}(real)  & 0.824±0.048          & 0.822±0.070          & 0.688±0.048          & 0.808±0.083           \\
                –mix 100\% & 0.824±0.055          & 0.838±0.066          & 0.708±0.039          & 0.836±0.054           \\
                –mix 200\% & 0.837±0.048          & 0.849±0.066          & 0.718±0.054          & 0.855±0.052           \\
                –mix 300\% & 0.837±0.038          & 0.842±0.064          & 0.735±0.034          & 0.849±0.049           \\
                –mix 400\% & \textcolor{red}{0.856±0.027} & 0.864±0.064          & \textcolor{red}{0.768±0.020} & 0.867±0.055           \\
                –mix 500\% & 0.841±0.038          & \textcolor{red}{0.869±0.055} & 0.757±0.042          & \textcolor{red}{0.872±0.043}  \\
            \bottomrule
            \multicolumn{4}{l}{\small \(^b\) means binary classification. \(^f\) means four-class classification}\\
            \end{tabular}
        }
    \end{minipage}\hfill
    \begin{minipage}{0.40\textwidth}
        \centering
        \includegraphics[width=\textwidth]{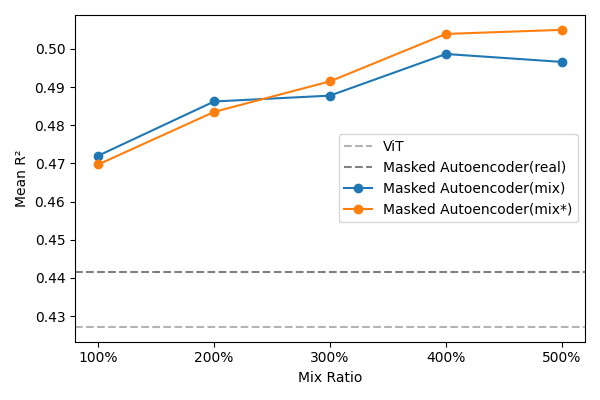}
        \caption{Performance comparison of mean \(R^2\) accross 82 cardiac phenotypes.} 
        \label{fig3}
    \end{minipage}
\end{figure}

\section{Conclusion}
In this study, we propose a novel two stage framework for CMR synthesis. To address the challenge of accurately generating the intricate structural and functional details of the heart, we introduce a cardiac phenotypes generation model at the first stage. By conditioning on these phenotypes, our approach efficiently generates a large volume of high-fidelity CMR cine sequences at stage two to augment pretraining data. Extensive experiments conducted on the large-scale, publicly available UKB dataset, as well as a private cardiac disease dataset, validate the superiority of our method over state-of-the-art models in terms of image synthesis, model pretraining and clinical applications.

\bibliographystyle{splncs04}
\bibliography{ref}

\begin{thebibliography}{10}
\providecommand{\url}[1]{\texttt{#1}}
\providecommand{\urlprefix}{URL }
\providecommand{\doi}[1]{https://doi.org/#1}

\bibitem{al2023usability}
Al~Khalil, Y., Amirrajab, S., Lorenz, C., Weese, J., Pluim, J., Breeuwer, M.: On the usability of synthetic data for improving the robustness of deep learning-based segmentation of cardiac magnetic resonance images. Medical Image Analysis  \textbf{84},  102688 (2023)

\bibitem{alexey2020image}
Alexey, D.: An image is worth 16x16 words: Transformers for image recognition at scale. arXiv preprint arXiv: 2010.11929  (2020)

\bibitem{bai2020population}
Bai, W., Suzuki, H., Huang, J., Francis, C., Wang, S., Tarroni, G., Guitton, F., Aung, N., Fung, K., Petersen, S.E., et~al.: A population-based phenome-wide association study of cardiac and aortic structure and function. Nature medicine  \textbf{26}(10),  1654--1662 (2020)

\bibitem{bluethgen2024vision}
Bluethgen, C., Chambon, P., Delbrouck, J.B., van~der Sluijs, R., Po{\l}acin, M., Zambrano~Chaves, J.M., Abraham, T.M., Purohit, S., Langlotz, C.P., Chaudhari, A.S.: A vision--language foundation model for the generation of realistic chest x-ray images. Nature Biomedical Engineering pp. 1--13 (2024)

\bibitem{chen2024ultrasound}
Chen, T., Shi, Y., Zheng, Z., Yan, B., Hu, J., Zhu, X.X., Mou, L.: Ultrasound image-to-video synthesis via latent dynamic diffusion models. In: International Conference on Medical Image Computing and Computer-Assisted Intervention. pp. 764--774. Springer (2024)

\bibitem{ding2024cross}
Ding, Z., Hu, Y., Li, Z., Zhang, H., Wu, F., Xiang, Y., Li, T., Liu, Z., Chu, X., Huang, Z.: Cross-modality cardiac insight transfer: A contrastive learning approach to enrich ecg with cmr features. In: International Conference on Medical Image Computing and Computer-Assisted Intervention. pp. 109--119. Springer (2024)

\bibitem{he2022masked}
He, K., Chen, X., Xie, S., Li, Y., Doll{\'a}r, P., Girshick, R.: Masked autoencoders are scalable vision learners. In: Proceedings of the IEEE/CVF conference on computer vision and pattern recognition. pp. 16000--16009 (2022)

\bibitem{heusel2017gans}
Heusel, M., Ramsauer, H., Unterthiner, T., Nessler, B., Hochreiter, S.: Gans trained by a two time-scale update rule converge to a local nash equilibrium. Advances in neural information processing systems  \textbf{30} (2017)

\bibitem{hou2023diversity}
Hou, Z., Yan, R., Wang, Q., Lang, N., Zhou, X.: Diversity-preserving chest radiographs generation from reports in one stage. In: International Conference on Medical Image Computing and Computer-Assisted Intervention. pp. 482--492. Springer (2023)

\bibitem{kingma2013auto}
Kingma, D.P.: Auto-encoding variational bayes. arXiv preprint arXiv:1312.6114  (2013)

\bibitem{li2023mage}
Li, T., Chang, H., Mishra, S., Zhang, H., Katabi, D., Krishnan, D.: Mage: Masked generative encoder to unify representation learning and image synthesis. In: Proceedings of the IEEE/CVF Conference on Computer Vision and Pattern Recognition. pp. 2142--2152 (2023)

\bibitem{li2024autoregressive}
Li, T., Tian, Y., Li, H., Deng, M., He, K.: Autoregressive image generation without vector quantization. arXiv preprint arXiv:2406.11838  (2024)

\bibitem{reynaud2023feature}
Reynaud, H., Qiao, M., Dombrowski, M., Day, T., Razavi, R., Gomez, A., Leeson, P., Kainz, B.: Feature-conditioned cascaded video diffusion models for precise echocardiogram synthesis. In: International Conference on Medical Image Computing and Computer-Assisted Intervention. pp. 142--152. Springer (2023)

\bibitem{rombach2022high}
Rombach, R., Blattmann, A., Lorenz, D., Esser, P., Ommer, B.: High-resolution image synthesis with latent diffusion models. In: Proceedings of the IEEE/CVF conference on computer vision and pattern recognition. pp. 10684--10695 (2022)

\bibitem{shen2023cellgan}
Shen, Z., Cao, M., Wang, S., Zhang, L., Wang, Q.: Cellgan: Conditional cervical cell synthesis for augmenting cytopathological image classification. In: International Conference on Medical Image Computing and Computer-Assisted Intervention. pp. 487--496. Springer (2023)

\bibitem{tudosiu2024realistic}
Tudosiu, P.D., Pinaya, W.H., Ferreira Da~Costa, P., Dafflon, J., Patel, A., Borges, P., Fernandez, V., Graham, M.S., Gray, R.J., Nachev, P., et~al.: Realistic morphology-preserving generative modelling of the brain. Nature Machine Intelligence  \textbf{6}(7),  811--819 (2024)

\bibitem{unterthiner2019fvd}
Unterthiner, T., van Steenkiste, S., Kurach, K., Marinier, R., Michalski, M., Gelly, S.: Fvd: A new metric for video generation  (2019)

\bibitem{wang2024self}
Wang, J., Wang, K., Yu, Y., Lu, Y., Xiao, W., Sun, Z., Liu, F., Zou, Z., Gao, Y., Yang, L., et~al.: Self-improving generative foundation model for synthetic medical image generation and clinical applications. Nature Medicine pp.~1--9 (2024)

\bibitem{wang2023modelscope}
Wang, J., Yuan, H., Chen, D., Zhang, Y., Wang, X., Zhang, S.: Modelscope text-to-video technical report. arXiv preprint arXiv:2308.06571  (2023)

\bibitem{wang2024screening}
Wang, Y.R., Yang, K., Wen, Y., Wang, P., Hu, Y., Lai, Y., Wang, Y., Zhao, K., Tang, S., Zhang, A., et~al.: Screening and diagnosis of cardiovascular disease using artificial intelligence-enabled cardiac magnetic resonance imaging. Nature Medicine  \textbf{30}(5),  1471--1480 (2024)

\bibitem{yan2021videogpt}
Yan, W., Zhang, Y., Abbeel, P., Srinivas, A.: Videogpt: Video generation using vq-vae and transformers. arXiv preprint arXiv:2104.10157  (2021)

\bibitem{ye2023synthetic}
Ye, J., Ni, H., Jin, P., Huang, S.X., Xue, Y.: Synthetic augmentation with large-scale unconditional pre-training. In: International Conference on Medical Image Computing and Computer-Assisted Intervention. pp. 754--764. Springer (2023)

\bibitem{yu2023language}
Yu, L., Lezama, J., Gundavarapu, N.B., Versari, L., Sohn, K., Minnen, D., Cheng, Y., Birodkar, V., Gupta, A., Gu, X., et~al.: Language model beats diffusion--tokenizer is key to visual generation. arXiv preprint arXiv:2310.05737  (2023)

\bibitem{zakeri2023dragnet}
Zakeri, A., Hokmabadi, A., Bi, N., Wijesinghe, I., Nix, M.G., Petersen, S.E., Frangi, A.F., Taylor, Z.A., Gooya, A.: Dragnet: learning-based deformable registration for realistic cardiac mr sequence generation from a single frame. Medical Image Analysis  \textbf{83},  102678 (2023)

\bibitem{zhao2023label}
Zhao, Z., Yang, J., Faghihroohi, S., Huang, K., Maier, M., Navab, N., Nasseri, M.A.: Label-preserving data augmentation in latent space for diabetic retinopathy recognition. In: International Conference on Medical Image Computing and Computer-Assisted Intervention. pp. 284--294. Springer (2023)

\bibitem{zhou2024heartbeat}
Zhou, X., Huang, Y., Xue, W., Dou, H., Cheng, J., Zhou, H., Ni, D.: Heartbeat: Towards controllable echocardiography video synthesis with multimodal conditions-guided diffusion models. In: International Conference on Medical Image Computing and Computer-Assisted Intervention. pp. 361--371. Springer (2024)

\end{thebibliography}
\end{document}